\documentclass[letterpaper, 10 pt, conference]{ieeeconf}  

\IEEEoverridecommandlockouts
\usepackage{cite}
\usepackage{amsmath,amssymb,amsfonts}
\usepackage{algorithmicx}
\usepackage{algorithm}
\usepackage{algpseudocode}
\algrenewcommand\algorithmicrequire{\textbf{Input:}}
\algrenewcommand\algorithmicensure{\textbf{Output:}}
\usepackage{graphicx}
\usepackage{textcomp}
\usepackage{xcolor}
\usepackage{pgfplots}
\newcommand\Rb{\mathbb{R}}

\usepackage{csquotes}

\usepackage{tikz}
\usepackage{pgfplots}
\usepackage{xcolor}
\usepackage{subcaption}
\pgfplotsset{compat=1.18}
\usetikzlibrary{positioning}
\usetikzlibrary{decorations.pathreplacing}
\usetikzlibrary{calc}
\usetikzlibrary{positioning,shapes,shadows,arrows}
\usepackage[font=small]{caption}

\pgfplotsset{
every axis/.append style={
  axis line style={->}, 
  legend style={font=\scriptsize},
  label style={font=\scriptsize},
  title style={font=\scriptsize},
  tick label style={font=\scriptsize},
  }
}

\makeatletter
\let\NAT@parse\undefined
\makeatother
\usepackage{hyperref}

\title{\LARGE \bf Non-Parametric Self-Identification and Model Predictive Control of Dexterous In-Hand Manipulation}
    
\author{%
    Podshara Chanrungmaneekul$^1$, 
    Kejia Ren$^1$,
    Joshua T. Grace$^2$,
    Aaron M. Dollar$^2$,
    and Kaiyu Hang$^1$
    \thanks{$^1$Department of Computer Science, Rice University, Houston, TX 77005, USA. \{\tt\small pc45, kr43, kaiyu.hang\}@rice.edu.}
    \thanks{$^2$Department of Mechanical Engineering and Material Science, Yale University, New Haven, CT 06511, USA. \{\tt\small josh.grace, aaron.dollar\}@yale.edu.}
    \thanks{This work was supported by the US National Science Foundation grant FRR-2133110 and FRR-2132823.}
}
\begin{document}
\maketitle

\begin{abstract}

    Building hand-object models for dexterous in-hand manipulation remains a crucial and open problem. Major challenges include the difficulty of obtaining the geometric and dynamical models of the hand, object, and time-varying contacts, as well as the inevitable physical and perception uncertainties. 
    Instead of building accurate models to map between the actuation inputs and the object motions, this work proposes to enable the hand-object systems to continuously approximate their local models via a self-identification process where an underlying manipulation model is estimated through a small number of exploratory actions and non-parametric learning. With a very small number of data points, as opposed to most data-driven methods, our system self-identifies the underlying manipulation models online through exploratory actions and non-parametric learning. By integrating the self-identified hand-object model into a model predictive control framework, the proposed system closes the control loop to provide high accuracy in-hand manipulation. Furthermore, the proposed self-identification is able to adaptively trigger online updates through additional exploratory actions, as soon as the self-identified local models render large discrepancies against the observed manipulation outcomes.
    We implemented the proposed approach on a sensorless underactuated Yale Model O hand with a single external camera to observe the object's motion. With extensive experiments, we show that the proposed self-identification approach can enable accurate and robust dexterous manipulation without requiring an accurate system model nor a large amount of data for offline training.
\end{abstract}

\section{Introduction}

Dexterous in-hand manipulation is a system-level problem consisting of an array of sub-problems, ranging from the modeling of contacts and hand-object dynamics \cite{Okamura2000, kao2016contact}, to the perception, planning, and control of the task-oriented hand-object coordination \cite{hang2016hierarchical, Kumar16, yuan2017gelsight}. At the core of all these problems, almost all existing approaches are challenged by the gaps between the required prior knowledge and online feedback, and the actual limited information available to the system \cite{Butepage19}. For example, a very common assumption made to contact-based manipulation systems is that the object model is perfectly known. In practice, however, this is rarely possible even if many sensing modalities are available. As such, in-hand manipulation systems are often limited either in their capability of handling complex dynamics or in their generalizability across similar variations of task setups. Although learning-based approaches have been extensively investigated and shown the capabilities of acquiring complex manipulation skills \cite{levine2016learning, zeng2018learning}, the data, which is the enabling factor for such systems, is also a major limitation in more general and dynamic tasks.


\begin{figure}[!t]    
\centerline{
    \includegraphics[width=\linewidth]{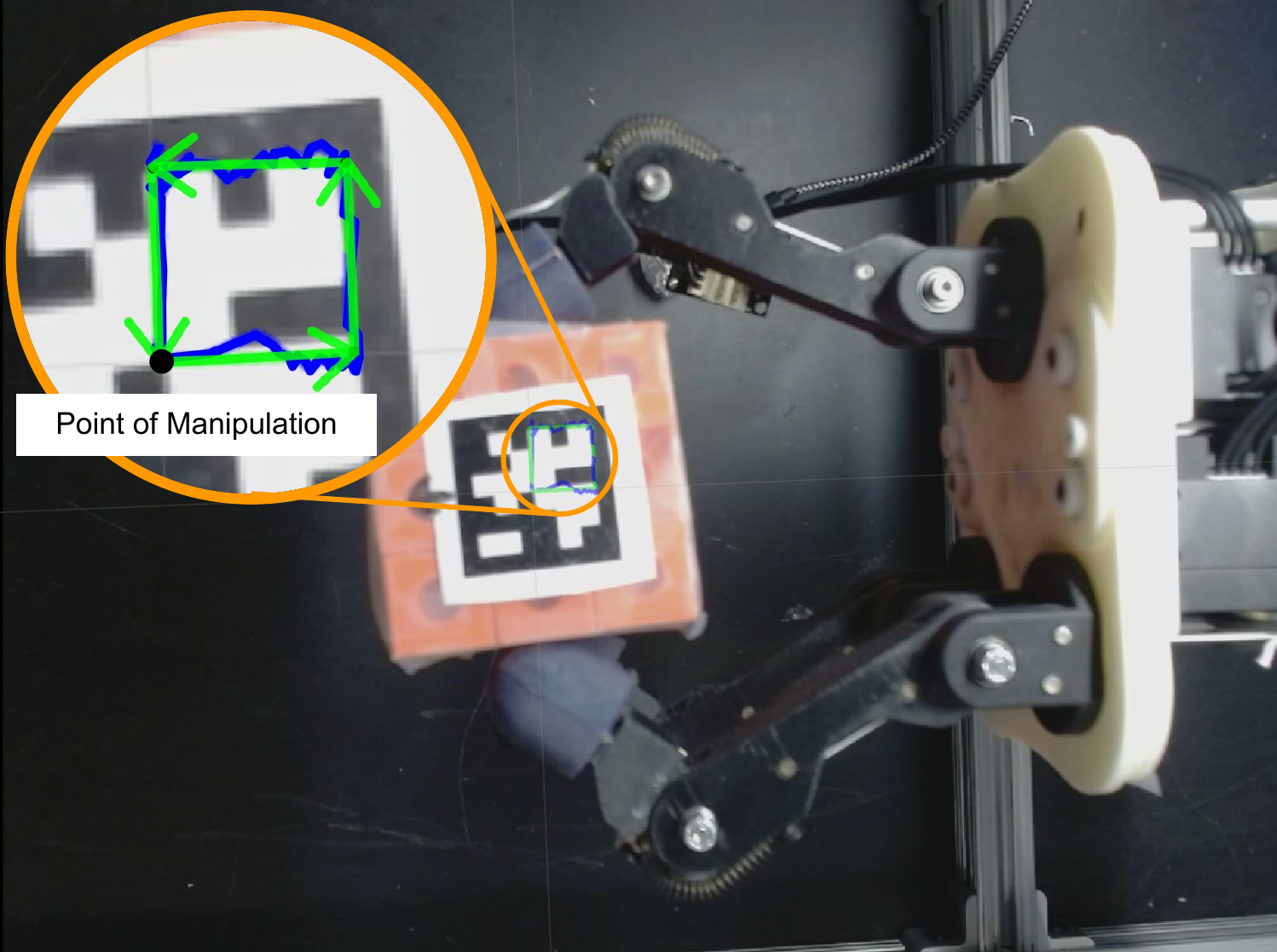}
}
\caption{An encoderless underactuated Yale Model O hand is tasked to manipulate an unknown object to trace a reference path (green trajectory). Enabled by the proposed non-parametric self-identification, and with no prior knowledge assumed, the hand is able to self-identify a system model using only $15$ data points and then accurately complete the task (blue trajectory). }
\label{fig:overview}
\vspace{-0.3cm}
\end{figure}

To bridge the aforementioned gaps while not shifting more burden to the sensing or data collection sides, we previously proposed the idea of \emph{self-identification}\cite{hang2021a} . For hand-object systems modeled with a number of known and missing parameters, the missing ones were iteratively self-identified by the hand-object system through exploratory actions without adding any additional sensors. The self-identified system then showed great performance in precise dexterous manipulation while tracking the real-time changes of the missing parameters. This approach was inspired by human manipulation where we do not have accurate models of everything \emph{a priori}. Rather, humans often use a strategy that shifts the system paradigm from \enquote{sense, plan, and act} to \enquote{ act, sense, and plan}. However, \cite{hang2021a} still modeled the hand-object system analytically and required a number of parameters to be self-identified and tracked in real-time, rendering the approach not easily generalizable and computationally very expensive. 

\begin{figure}[!t]    
\centerline{
    \centerline{\includegraphics[width=0.95\linewidth]{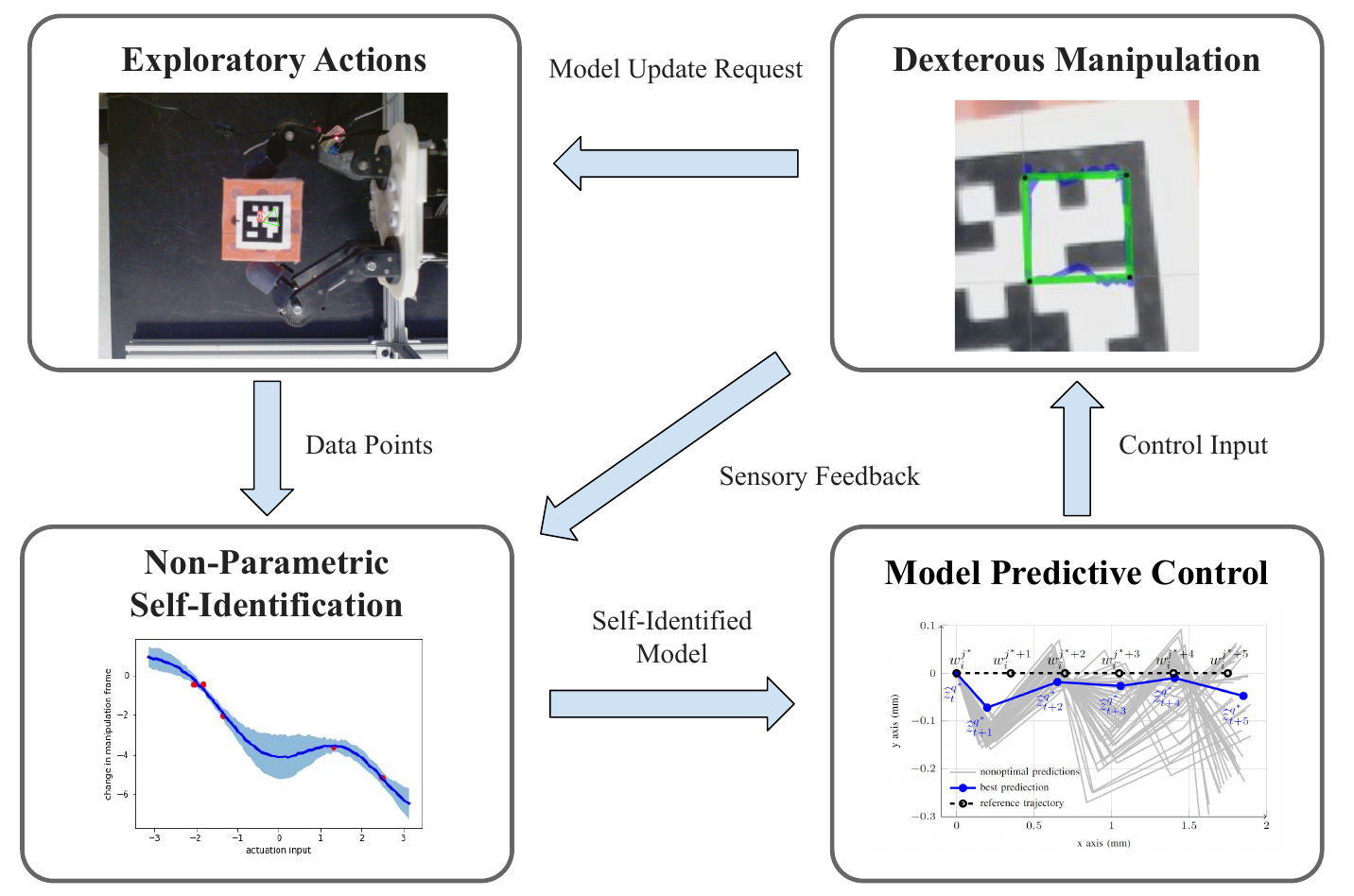}}
}
\caption{System diagram of the proposed non-parametric self-identification for dexterous in-hand manipulation.}
\label{fig:diagram}
\vspace{-0.3cm}
\end{figure}

To this end, this work proposes to replace the analytic parameter-based models with a non-parametric model to be self-identified. We consider a challenging setup with an encoderless underactuated robot hand, as shown in Fig.~\ref{fig:overview}. Given an unknown grasp on an unknown object, the proposed system first collects a small number of manipulation data points through exploratory actions, for which the grasp stability is passively secured by the hand's compliance. The system then learns a non-parametric model to map from the hand control directly to the object motion, yielding a self-identified local model of the hand-object system. In this work, the non-parametric model was learned by a Gaussian Process Regressor. By integrating the self-identified model into a Model Predictive Control (MPC) framework, we show that in-hand manipulation can be precisely achieved, while the self-identified model can be updated through additional exploratory actions as needed. A system diagram is illustrated in Fig.~\ref{fig:diagram}.

\section{Related work}
\label{related work}

\emph{Hand-Object Models:} Traditional models of in-hand manipulation systems often assume that precise geometric and physical models of the hand, object, contacts, etc., are available \cite{kao2016contact}. Since such approaches are very sophisticated in modeling every detailed aspect of the system, they are limited in scalability and normally focus on specific sub-problems of hand-object systems, including contact modeling \cite{sundaralingam2018geometric}, force control \cite{li2018learning}, stability maintenance \cite{hang2016hierarchical}. More importantly, model-based methods are inherently limited as the assumptions of model availability often do not hold in reality. On another hand, simplified models such as action primitives have been designed to model the manipulation mappings \cite{Calli19, liarokapis2016post2}. However, as the primitives are handcrafted, they are not generalizable, nor scalable. This work proposes self-identifying a non-parametric model of the hand-object system, aiming at avoiding sophisticated modeling, model availability assumption, and unnecessary model simplification.

\emph{Data-Driven Approaches:} With sufficient data and training, learning-based methods have shown unprecedented performance in acquiring complex manipulation skills \cite{zeng2018learning}. In an end-to-end manner, data has filled in the gap traditionally formed by the lack of \emph{a priori} system information and perception uncertainties \cite{levine2016learning, kaelbling2020foundation}. However, as such methods are sensitive to the amount and diversity of the training data, they are often not generalizable, even to similar task variations. Unlike those data-demanding approaches, the non-parametric Gaussian Process model employed in this work is lightweight and known to work with a minimal amount of data \cite{fazeli2017learning}. As such, it enables the self-identification of hand-object models online through only a few exploratory actions.

\emph{Interactive Perception:} Leveraging proactive manipulation actions to unveil hidden system information can greatly improve the robot perception under limited sensing \cite{bohg17}. Particularly for hand-object systems, interactive perception can enable grasping and in-hand manipulation under large uncertainties \cite{koval2015pose, platt2017efficient}. While most interactive perception methods focus on estimating specific parameters of a system, our non-parametric self-identification aims to directly build a mapping, approximated locally, from the hand's actuation input to the object motions to enable the MPC control of precise dexterous manipulation.

\section{Hand-object systems and problem formulation}
\label{problem}

In this work, we aim to address the in-hand dexterous manipulation problem, where an object grasped by an underactuated robot hand needs to be reconfigured to certain poses.
We consider the hand and the object as a whole discrete-time dynamical system.
The underlying state of the hand-object system at time $t$ is $s_t \in \Rb^N$, where $N$ is the number of all the physical properties necessary for uniquely identifying a system state, such as hand joint configurations and hand-object contact locations.
The control of the system, $u_t \in \Rb^C$, is the actuation input to the hand at time $t$.
For an underactuated hand, the dimension of controls, $C$, is less than the degree of freedom of the hand.
The dynamics of the system can be represented by a transition function $g: \Rb^N \times \Rb^C \mapsto \Rb^N$ such that
\begin{equation}
    s_{t+1} = g(s_t, u_t)
\end{equation}
Additionally, we select a fixed point on the object's surface and use this point's motion to represent the object's motion.
We term this point as the Point of Manipulation (POM), whose position at time $t$ is denoted by $z_t \in \Rb^3$.
As such, the object's motion at time $t$ can be represented by the finite difference of POM's positions, i.e., $\delta z_t = z_{t+1} - z_t$.

However, building an analytical model for such hand-object systems is impossible due to the following challenges: 1) the system state $s_t$ is not fully observable as it contains parameters not obtainable due to the limited sensing capability of the system, such as the hand-object contact locations and the joint angles of the underactuated hand; and 2) the system dynamics $g$ requires accurate geometric models of the hand and the object, which are in general unavailable.
Moreover, even if $g$ is solvable, it is hard to generalize across different object shapes or different types of contacts.
Instead, assuming a neglectable change of the hand-object system state after applying a small enough control, we can locally approximate the system model without exactly knowing the current system state $s_t$.
For this, we use another function $\Gamma: \Rb^C \to \Rb^3$ to represent the locally approximated system transitions, which maps from the control input to the object's motion:
\begin{equation}
    \Gamma(u_t) = \delta z_t
\label{eq:Gamma}
\end{equation}
In addition, the inverse of the approximated system model is defined by $\Gamma^{-1}: \Rb^3 \to \Rb^C$.
We name $\Gamma$ and its inverse $\Gamma^{-1}$ as \emph{local manipulation models}.
To precisely manipulate the object without prior knowledge of the system dynamics, the system needs to self-identify the local manipulation models $\Gamma$ and $\Gamma^{-1}$ and adapt them to different hand-object configurations when necessary.

In this work, the hand-object system is tasked to find and execute a sequence of control inputs to gradually move the object, such that the POM will trace through a reference trajectory represented by a sequence of $T$ desired positions of POM: $X = \{x_1, \cdots, x_T\}$, where $x_1, \cdots, x_T \in \Rb^3$ are called \emph{keypoints} of the trajectory.
We formulate such dexterous in-hand manipulation as a self-identification and control problem, and approach the problem through non-parametric learning and Model Predictive Control (MPC), as illustrated in Fig.~\ref{fig:diagram} and summarized in Alg.~\ref{alg:control-system}.
The details of the approach will be described in Sec.~\ref{GP} and Sec.~\ref{mpc}.

Starting with the POM positioned at $z_0 \in \Rb^3$, the system identifies the models $\Gamma$ and $\Gamma^{-1}$ through a small number of initial exploratory actions
consisting of $d$ randomly sampled and $a$ calculated controls,
as will be detailed in Sec.~\ref{actions} and Alg.~\ref{alg:action-selection}.
Then, the self-identified models will be integrated into an MPC framework to generate real-time control $u_t$ to move POM toward targeted keypoints of the reference trajectory.
Meanwhile, the system observes the outcome position of POM $z_{t+1}$ after each generated control $u_t$ has been executed.
If a large deviation from the desired reference trajectory has been detected, as will be described in Sec.~\ref{resample},
the system will update the models $\Gamma$ and $\Gamma^{-1}$ by performing $b$ more exploratory actions.
As such, being self-identified and updated in real-time, the models are used to generate controls to precisely move the POM on the object to reach each keypoint of the reference trajectory sequentially.

 
\begin{algorithm}[t]
    \caption{Dexterous Manipulation via Self-Id and MPC}
    \label{alg:control-system}
    \small
    \begin{algorithmic}[1]
    \Require Reference trajectory $X$, a distance threshold $\alpha$, number of initial exploratory actions $d$ and $a$, number of adapting actions for model update $b$
        \State $t \gets 0$, $z_0 \gets \Call{ObservePOM()}{}$
        \State $x_0 \gets z_0$
        \State $\Gamma, \Gamma^{-1} \gets \Call{SelfIdentification}{z_0, d, a}$ \hfill \Comment{Alg.~\ref{alg:action-selection}}
        \For{$x_i \in X$, $i = 1, \cdots, n$} \hfill \Comment{Waypoints in $X$}
            \While{$\|x_i - z_t \| > \alpha$}
                \State $u_t \gets \Call{MPC}{z_t, x_{i-1}, x_i}$ \hfill \Comment{Alg.~\ref{alg:mpc}}
                \State $z_{t+1} \gets \Call{Execute}{u_t}$ \hfill \Comment{Observe POM}
                \If{$\epsilon_t > \gamma$} \hfill \Comment{Sec.~\ref{resample}}
                    \State $\Gamma, \Gamma^{-1} \gets \Call{SelfIdentification}{z_{t+1}, 0, b}$
                \EndIf
                \State $t \gets t+1$
            \EndWhile
        \EndFor
    \end{algorithmic}
\end{algorithm}
\vspace{-.1cm}
\section{Non-Parametric Self-Identification}
\label{GP}



In this section, we present a non-parametric approach based on Gaussian Process Regression to facilitate the self-identification of the local manipulation models $\Gamma$ and $\Gamma^{-1}$.
Such a non-parametric learning approach does not require a parametric form of the models, which is challenging to specify and difficult to generalize.
Moreover, as an efficient nonlinear function approximator that works well with a small amount of data, Gaussian Process Regression alleviates the burden of heavy online data collection, which is time-consuming for a real-world system.
Specifically, as described in Sec.~\ref{actions}, with a set of data points collected by the system through online exploratory actions, the manipulation models $\Gamma$ and $\Gamma^{-1}$ can be learned efficiently to find the inherent relation between the control inputs and the object's motion.


\subsection{Exploratory Actions and Self-Identification}
\label{actions}
To self-identify the manipulation models with data collected online,
we dynamically maintain a training dataset, $\mathcal{D} = \{(\hat{u}_i, \delta \hat{z}_i)\}_{i=1}^P$, consisting of $P$ data points the system has observed.
Each data point is a pair $(\hat{u}_i, \delta \hat{z}_i)$, where $\hat{u}_i \in \Rb^C$ is a control the system has executed and $\delta \hat{z}_i \in \Rb^3$ is the object's motion observed after executing $\hat{u}_i$.
The dataset $\mathcal{D}$ is initially empty but will be updated to have more data points as the system keeps executing to manipulate the object.
We use \emph{exploratory actions} to name the data points in the training dataset $\mathcal{D}$, as such actions are performed for exploring the system models.
We illustrate how such exploratory actions are generated and used for self-identification in Alg.~\ref{alg:action-selection}.


Without prior knowledge about the hand-object configuration, the system begins by randomly generating a number of $d$ controls. 
Each control is randomly sampled from a $C$-dimensional uniform distribution within the range $[-l, l]$, where $l$ is chosen to be arbitrarily small while not being overwhelmed by the system's physical uncertainties. 
The system will execute each of these $d$ controls, observe the object's motion after each control execution, and add them to the training dataset $\mathcal{D}$.


However, certain patterns of the object's motions might not be present in $\mathcal{D}$ since the size $P$ of the dataset is kept small in practice.
Therefore, to have more representative data to effectively learn the manipulation models, we intend to increase the local density of the dataset $\mathcal{D}$ by selecting additional $a$ controls to explore.
For that, we define the local density of the dataset $\mathcal{D}$ at its $i$-th data point to be the reciprocal of the distance between $\delta \hat{z}_i$ and its nearest neighbor in $\mathcal{D}$:
\begin{equation}
    \rho_\mathcal{D}(i) = \frac{1}{\min_{j \neq i}\lVert \delta \hat{z}_i - \delta \hat{z}_j \rVert}
\end{equation}
The data point with the lowest local density will be used to calculate a new control $\hat{u}_s$, to be added into the training dataset $\mathcal{D}$ with its corresponding observation of the object's motion $\delta \hat{z}_s$.
This new control $\hat{u}_s$ is determined by the average of this data point and its nearest neighbor:
\begin{equation}
\label{eq:us}
\begin{aligned}
    \hat{u}_s &= \frac{\hat{u}_p + \hat{u}_{p'}}{2}\\
    p &= \underset{j \in \{1, \cdots, \lvert \mathcal{D} \rvert\}}{\arg\min} \rho_\mathcal{D} (j)\\
    p' &=  \underset{j \in \{1, \cdots, \lvert \mathcal{D} \rvert\} \backslash \{p\}}{\arg\min} \lVert \delta \hat{z}_p - \delta \hat{z}_j \rVert
\end{aligned}
\end{equation}
Using this method, we would approach a uniform distribution as the number of exploratory actions increases.
With the dataset $\mathcal{D}$ generated by exploratory actions, the manipulation models $\Gamma$ and $\Gamma^{-1}$ can be efficiently self-identified by Gaussian Process Regression (Alg.~\ref{alg:action-selection}).
It is worth noting that both models $\Gamma$ and $\Gamma^{-1}$ are regressed with the same dataset $\mathbf{D}$, but with a different domain and codomain of the data.
As $\Gamma$ and $\Gamma^{-1}$ are independently learned, we cannot guarantee a closed loop between them.
In other words, for the self-identified models, $\Gamma^{-1}(\Gamma (u)) \neq u$.


\begin{algorithm}[t]
    \caption{Self-Identification}
    \label{alg:action-selection}
    \small
    \begin{algorithmic}[1]
    \Require Observed POM $z_0$, number of random actions $d$, number of extra actions $a$
    \Ensure Non-parametric manipulation models $\Gamma$ and $\Gamma^{-1}$
        \State $\mathcal{D} \gets \Call{AcquireDataset()}{}$ \hfill \Comment{Training Dataset}
        \For{$i = 1, \cdots, d$} \hfill \Comment{$d$ Random Actions}
            \State $\hat{u}_i \gets \Call{Uniform}{-l, l}$ \hfill \Comment{Uniform Sampling from $[-l, l]$}
            \State $z_i \gets \Call{Execute}{\hat{u}_i}$ \hfill \Comment{Observe POM}
            \State $\delta \hat{z}_i \gets z_i - z_{i-1}$ \hfill \Comment{Object's Motion}
            \State $\mathcal{D} \gets \mathcal{D} \cup \{(\hat{u}_i, \delta \hat{z}_i)\}$
        \EndFor
        \For{$i = d+1, \cdots, d+a$} \hfill \Comment{$a$ Extra Actions}
            \State $p \gets \mathrm{\arg\min}_j \rho_\mathcal{D}(j)$
            \State $p' \gets \mathrm{\arg\min}_{j \neq p} \lVert \delta \hat{z}_p - \delta \hat{z}_j \rVert$ \hfill \Comment{Nearest Neighbor}
            \State $\hat{u}_s \gets (\hat{u}_p + \hat{u}_{p'})/ 2$ \hfill \Comment{Eq.~\eqref{eq:us}}
            \State $z_{i} \gets \Call{Execute}{\hat{u}_s}$ \hfill \Comment{Observe POM}
            \State $\delta \hat{z}_s \gets z_i - z_{i-1}$ \hfill \Comment{Object's Motion}
            \State $\mathcal{D} \gets \mathcal{D} \cup \{(\hat{u}_s, \delta \hat{z}_s)\}$
        \EndFor
        \State $\Gamma, \Gamma^{-1} \gets \Call{GPR}{\mathcal{D}}$ \hfill \Comment{Gaussian Process Regression}
        \State \Return $\Gamma$, $\Gamma^{-1}$
    \end{algorithmic}
\end{algorithm}




\subsection{Model Update}
\label{resample}

\begin{figure}[htbp]
    \vspace{-0.3cm}
    \centerline{\includegraphics[width=0.91\columnwidth]{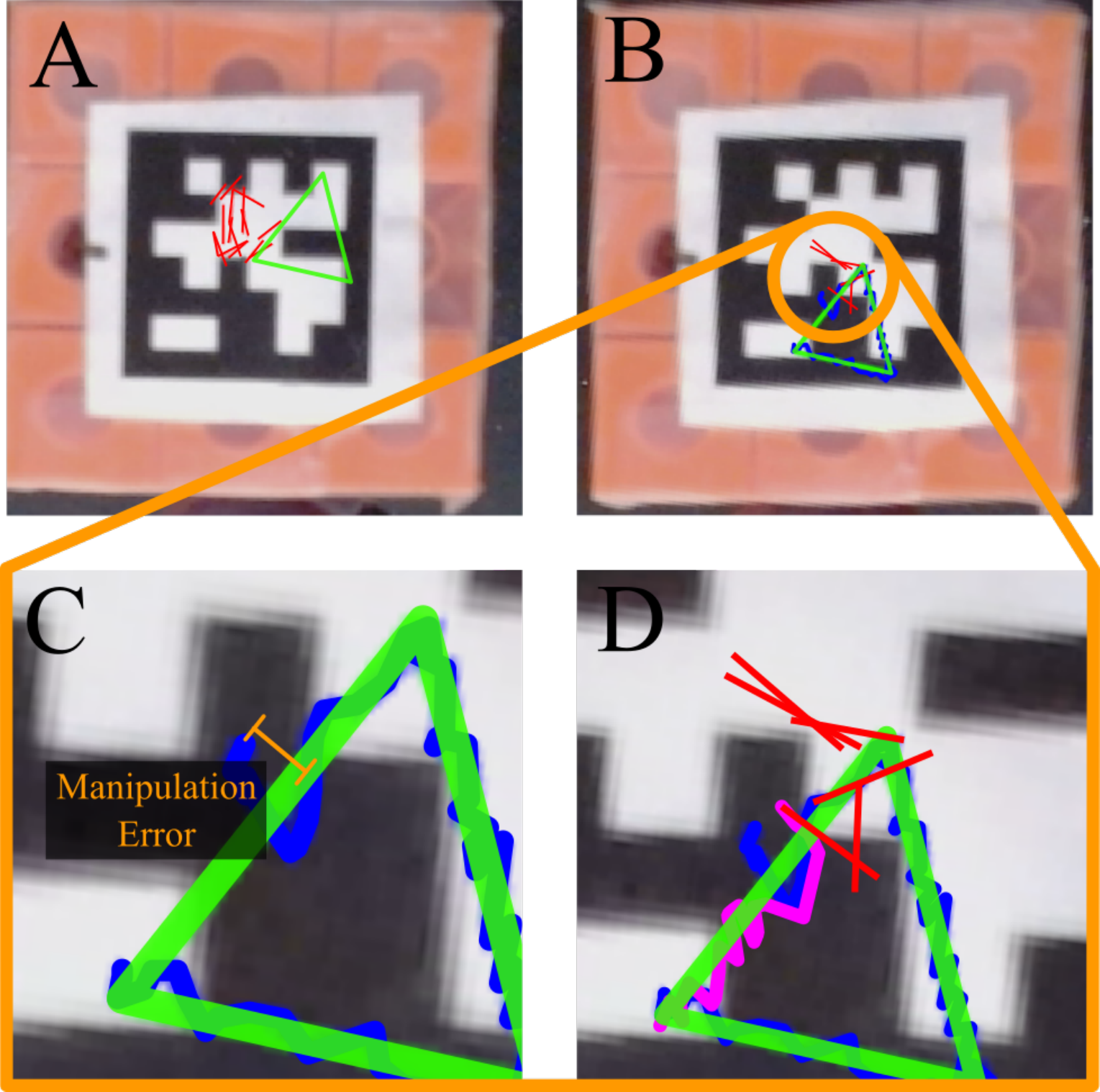}}
    \caption{An example demonstration of model update with the setup in Fig.~\ref{fig:overview}, where the system is tasked to trace the reference trajectory (green triangle): (A) At the beginning, the system performs $d+a$ initial exploratory actions (red) to learn the non-parametric models. (B) The model update request is triggered during manipulation, according to the sensory feedback. (C) The manipulation error $\epsilon_t$ (yellow) exceeds the threshold $\gamma$, triggering the model update. (D)  Additional $b$ data points obtained from adapting actions (red) are used to update the models. Then, the system continues with the trajectory tracing tasks (magenta). }
    \label{fig:resample}
    \vspace{-0.3cm}
\end{figure}

While manipulating the object, the underlying system dynamics can vary over time due to changes in hand configuration and contacts.
This can cause the failure of the self-identified models as they are locally approximated.
Therefore, we introduce a mechanism in our framework that adaptively updates the model online when needed.
As demonstrated with Fig.~\ref{fig:resample}, if a large discrepancy between the observed system state and the prediction of self-identified models has been detected, the model will be updated with $b$ additional actions calculated by Eq.~\eqref{eq:us}.
We particularly name such exploratory actions \emph{adapting actions}, as they are used to adapt the model to a new locality.

For that, we need to define an indicator, to determine when the model update should be triggered.
Consider the in-hand manipulation task defined in Sec.~\ref{problem}.
Suppose that the POM has already reached the first $i-1$ keypoints of the reference trajectory $X$ through in-hand manipulation.
In other words, the system currently targets the next keypoint $x_i \in X$.
We linearly interpolate from $x_{i-1}$ to $x_i$ to create an intermediate trajectory $W_i = \{w_i^1, w_i^2, \cdots, w_i^M\}$ of $M$ waypoints, where $w_i^1 = x_{i-1}$ and $w_i^M = x_i$.
Given POM's position $z_t \in \Rb^3$ at the current time step, the nearest waypoint $w_i^{j^*}$ in the intermediate trajectory $W_i$ is found by
\begin{equation} 
\label{eq:nearest_point}
    j^* = \underset{j \in \{1, \cdots, M\}}{\arg\min} \lVert z_t -  w_i^j \rVert
\end{equation}
With this, we define the manipulation error $\epsilon_t$ at time $t$ to be the distance between POM's position and its nearest waypoint in the intermediate trajectory:
\begin{equation} \label{eq:manipulation_error}
    \epsilon_t = \lVert z_t - w_i^{j^*} \rVert
\end{equation}
The manipulation error $\epsilon_t$ measures how much the POM deviates from its desired trajectory.
If $\epsilon_t$ is greater than a threshold $\gamma$, the framework will update the model.
This mechanism can be found in lines 8 and 9 in Alg.~\ref{alg:control-system}.


\subsection{Model Transfer}
\label{sec:transfer}
The underlying system dynamics become different when the object's geometry or grasp configuration has changed.
Intuitively, however, the patterns of manipulation can render some similarities across such geometric and physical variations.
In other words, the self-identified manipulation models should feature generalizability on different objects or different contact locations, facilitating model transfer between different hand-object setups.

As such, when manipulating a new object, our framework has the option to initialize the manipulation models $\Gamma$ and $\Gamma^{-1}$ by the models that have been learned previously with a different object.
By such model transfer, the system can skip the data-gathering step for initial exploratory actions in line 3 of Alg.~\ref{alg:control-system},
to speed up the manipulation of a new object.
We will show the benefits of using model transfer in our framework through the experiments in Sec.~\ref{ex3}.

\section{Model Predictive Control}
\label{mpc}
As the manipulation models $\Gamma$ and $\Gamma^{-1}$ can be efficiently self-identified through exploratory actions in Sec.~\ref{GP}, we can use them as predictive models to develop a control scheme to generate controls based on the desired motion of the object.
To this end, we integrate the models $\Gamma$ and $\Gamma^{-1}$ in a Model Predictive Control (MPC) framework to iteratively generate controls at each time step.
Our MPC-based control scheme is presented in Alg.~\ref{alg:mpc} and the details will be described below.
Benefitting from the efficient inference of $\Gamma$ and $\Gamma^{-1}$, the MPC effectively meets the requirement of real-time executions.

As some definitions in Sec.~\ref{resample} are useful for MPC, we briefly recall them here:
between the last reached keypoint $x_{i-1}$ and its next $x_i$ in the reference trajectory $X$, we create an intermediate trajectory $W_i = \{w_i^1, w_i^2, \cdots, w_i^M\}$ of $M$ waypoints by linear interpolation.
In this intermediate trajectory $W_i$, we find the nearest $w_i^{j^*}$, with its index $j^{*}$, to POM's position $z_t$ at the current time step.



\begin{figure}[htbp]
    {\input{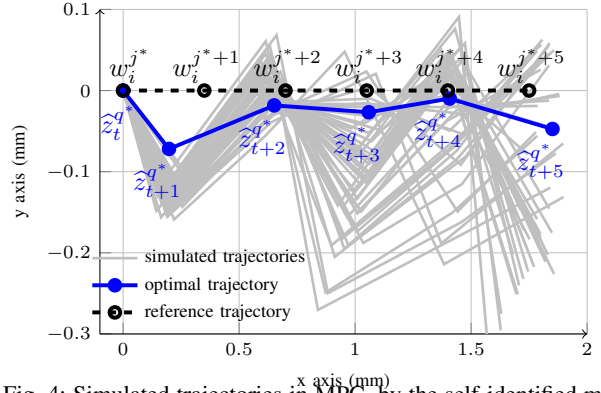}}
    \vspace{-0.4cm}
    \caption{Simulated trajectories in MPC, by the self-identified models $\Gamma$ and $\Gamma^{-1}$. In this figure, $Q=50$ trajectories (gray) are simulated and each one has a horizon of $L = 5$. The optimal trajectory (blue), closest to the reference trajectory (black), is selected to extract the optimal control.}
    \label{fig:mpc}
    \vspace{-0.3cm}
\end{figure}

Then, we use the intermediate trajectory $W_i$ as a local reference to guide the MPC in searching for optimal control.
Concretely, MPC uses the self-identified models $\Gamma$ and $\Gamma^{-1}$ to predict the behavior of the controlled hand-object system up to a prediction horizon $L$.
By adding stochasticity into the prediction with some random $\xi$ of control, it can simulate $Q$ independent trajectories, as illustrated in Fig.~\ref{fig:mpc}.
Each simulated trajectory $U^q = \{(\hat{u}_t^q, \hat{z}_t^q), \cdots, (\hat{u}_{t+L}^q, \hat{z}_{t+L}^q)\}$, where $q=1, \cdots, Q$, is generated by the following iterative process starting with $k=0$:



\begin{equation} 
\label{eq:prop}
\begin{aligned}  
    \hat{u}_{t+k}^q &= \Gamma^{-1}(w_i^{j^*+k} - \hat{z}_{t+k}^q) + \xi\\
    \hat{z}_{t+k+1}^q &= \Gamma(\hat{u}_{t+k}^q) + \hat{z}_{t+k}^q
\end{aligned}
\end{equation}
where $\hat{u}_{t_k}^q \in \Rb^C$ and $\hat{z}_{t+k}^q \in \Rb^3$ are the predicted control and state (i.e., POM's pose) at time $t+k$ in the $q$-th simulated trajectory, and $\xi \sim \mathcal{N}(0, \sigma \mathbb{I}_C)$ is a multivariate Gaussian random variable.
The scale $\sigma$ of this Gaussian random variable is named \emph{MPC optimization scale}.


Over the $Q$ simulated trajectories, MPC searches for the optimal trajectory $U^{q^*}$ (the blue one in Fig.~\ref{fig:mpc}) such that the accumulated distance between it and the intermediate trajectory $W_i$ is minimized:
\begin{equation}  
\label{eq:min}
    q^* = \underset{q\in\{1, \cdots, Q\}}{\arg\min}\left(\sum^L_{k=0}\|\hat{z}^q_{t+k} - w_i^{j^*+k}\|\right)
\end{equation}
where $L = \min\{K, M-j^*\}$ is the prediction horizon (i.e., the length of the simulated trajectories) not greater than a hyperparameter $K$.
The first control $\hat{u}_t^{q^*}$ in the optimal trajectory $U^{q^*}$ is then sent to the hand actuators for execution.
While the entire procedure of MPC is performed at each time step, the system will precisely control the object's motion, guided by the self-identified manipulation models.


\begin{algorithm}
    \caption{Model Predictive Control (MPC)}
    \label{alg:mpc}
    \small
    \begin{algorithmic}[1]
    \Require Observed POM $z_t$, last reached keypoint $x_{i-1}$, targeted keypoint $x_i$
    \Ensure Optimized control for execution $u_t$
        \State $\{w_i^1, \cdots, w_i^M\} \gets \Call{LinearInterpolate}{x_{i-1}, x_i}$
        \State $j^* \gets \arg\min_j \lVert z_t - w_i^j\Vert$ \hfill \Comment{Nearest Waypoint by Eq.~\eqref{eq:nearest_point}}
        \State $L \gets \min\{K, M - j^*\}$ \hfill \Comment{Prediction Horizon}
        \For{$q = 1, \cdots , Q$}
            \State $\hat{z}_t^q \gets z_t$
            \For{$k = 0, \cdots, L-1$}
                \State $\xi \gets \mathcal{N}(0, \sigma \mathbb{I}_C)$
                \State $\hat{u}_{t+k}^q \gets \Gamma^{-1}(w_i^{j^*+k} - \hat{z}_{t+k}^j) + \xi$ \hfill \Comment{Predicted Control}
                \State $\hat{z}_{t+k+1}^q \gets \Gamma(\hat{u}_{t+k}^q) + \hat{z}_{t+k}^q$ \hfill \Comment{Predicted State}
            \EndFor
        \EndFor
        \State $q^* \gets \arg\min_q \left(\sum^L_{k=0}\|\hat{z}^q_{t+k} - w_i^{j^*+k}\|\right)$ \hfill \Comment{Eq.~\eqref{eq:min}}
        \State \Return $u^{q^*}_t$
    \end{algorithmic}
\end{algorithm}

\section{Experimental Evaluation}
\label{experiment}
\begin{figure}[htbp]
    \vspace{-0.5cm}
    \centerline{\includegraphics[scale=0.25]{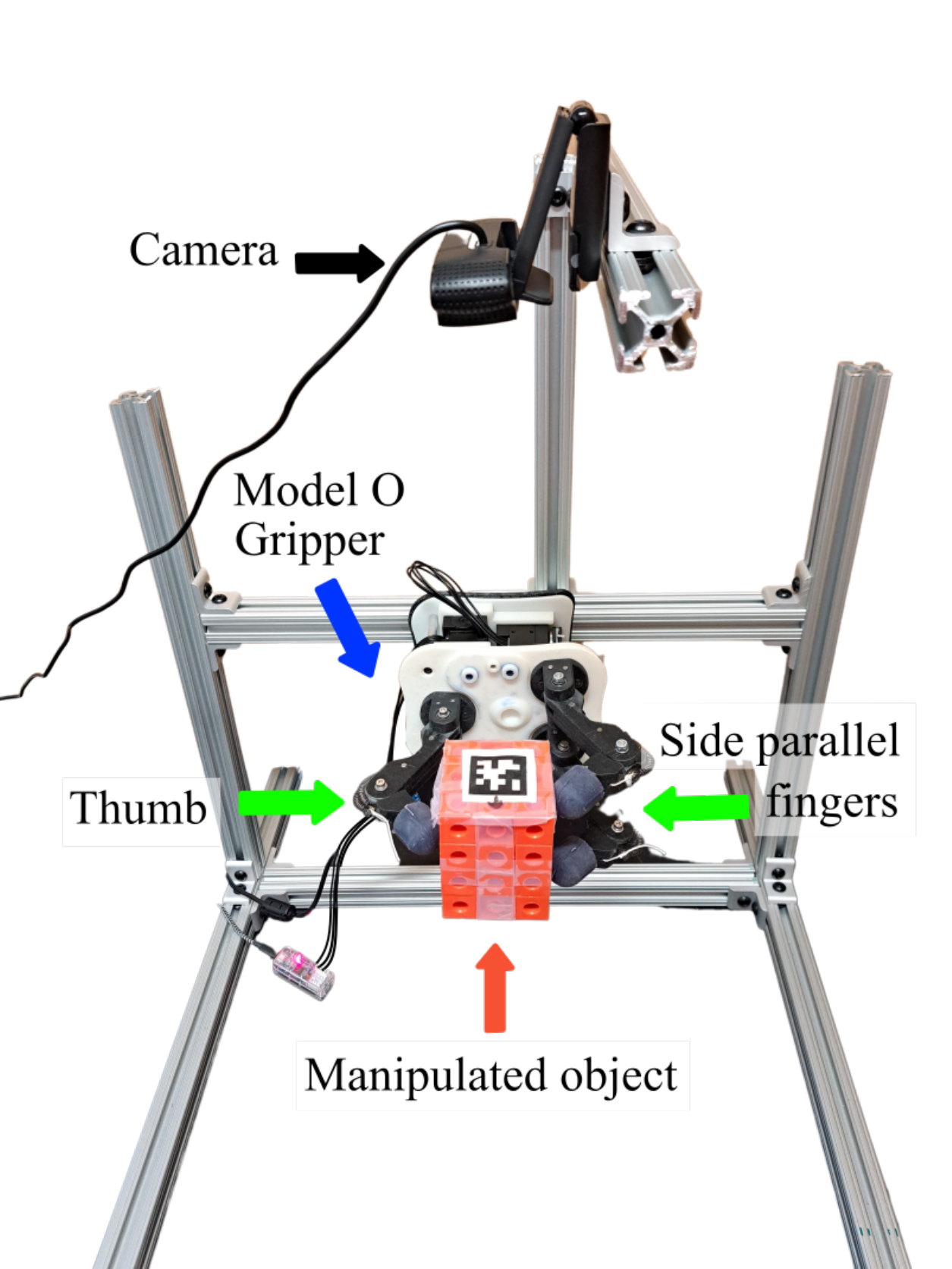}}
    \caption{The experimental setup: The Yale Model O underactuated hand is tasked to manipulate an object (an orange cube (obj \#4)), whose POM is tracked by a top camera with AprilTag.}
    \label{setup}
    \vspace{-0.3cm}
\end{figure}

In this section, we evaluate and study the performance of the proposed framework under a real-world setting that requires precise in-hand manipulation.
As shown in Fig.~\ref{setup}, We deployed our proposed framework on a Yale Model O underactuated hand\cite{odhner2014compliant}.
This hand has three identical fingers, each finger of which has one motor to actuate two spring-loaded joints through the tendon.
While the tendon is pulled by the motor, the joint configuration of each finger will change accordingly.
The spring in each finger joint enables compliance that facilitates stable contact between the hand and the grasped object.
For our experimental setup, we restricted two fingers to be parallel to each other and always take the same actuation input, while the third finger was configured to the opposite side.
As such, the object's motion was physically constrained in a horizontal plane.

\begin{figure}[htbp]
    \centerline{\includegraphics[height=5cm,keepaspectratio]{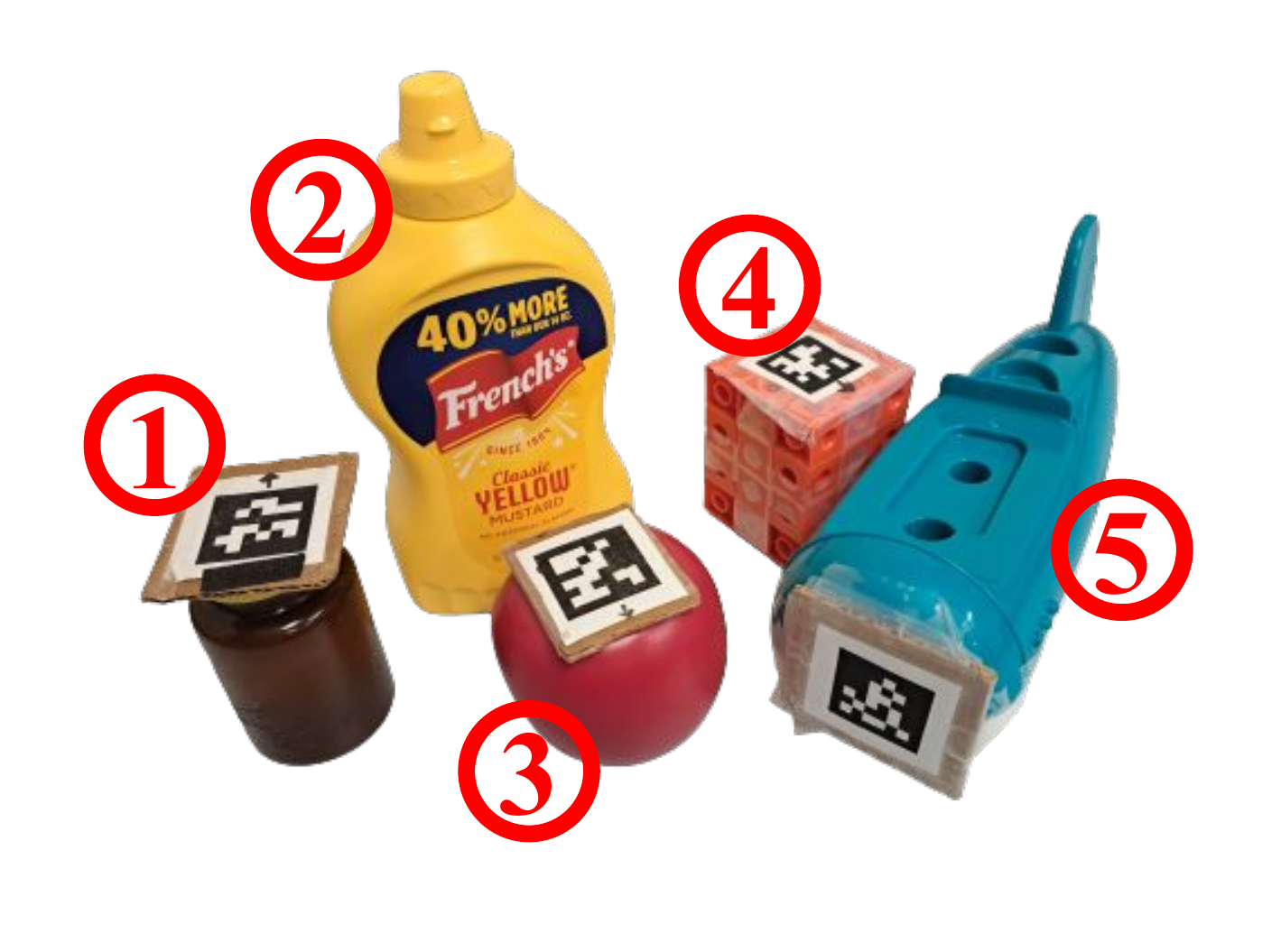}}
    \vspace{-0.5cm}
    \caption{The objects used in the experiments: 1) a pill bottle, 2) a mustard bottle (YCB dataset \#006), 3) an apple (YCB dataset \#013), 4) an orange cube, and 5) a toy airplane (YCB dataset \#072\-a)\cite{7251504}.}
    \label{object}
    \vspace{-0.3cm}
\end{figure}

The POM was selected to be on the top of the object, which was tracked by a camera mounted above the object through AprilTag\cite{5979561}.
The camera has a resolution of $1024 \times 512$ and the tracker's frequency is $30$ fps.
Note that the tag-based POM tracker can be replaced by other vision-based frameworks.
At the beginning of each experiment trial, the experimenter needed to hand one object in Fig.~\ref{object} to the underactuated hand by a stable grasp.

\subsection{Experiment Design}
\begin{figure*}[t]
    \centerline{\includegraphics[width=1.99\columnwidth]{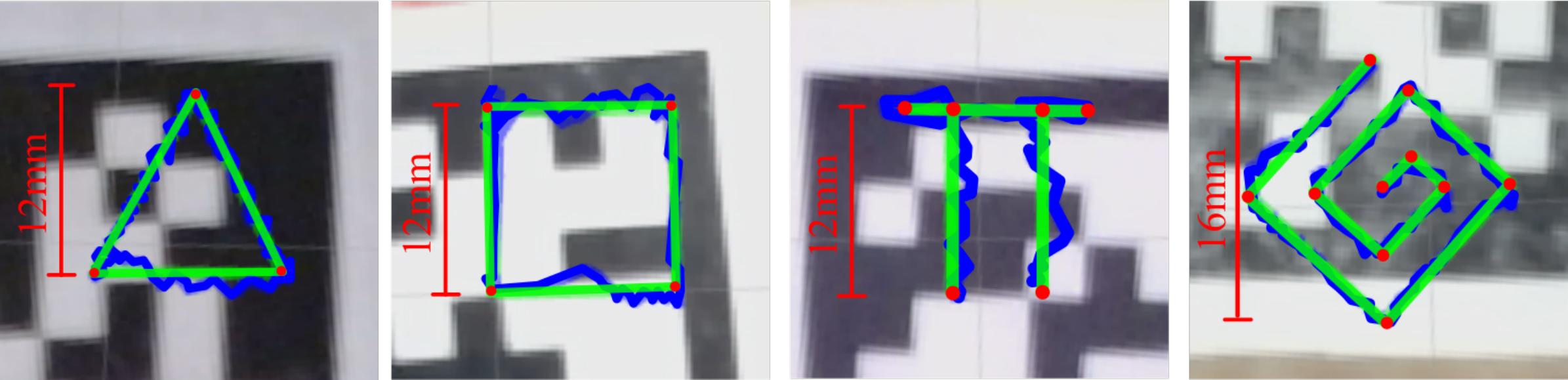}}
    \caption{Real-world trajectory tracing tasks by in-hand manipulation: a triangle, a square, a $\pi$ letter, and a spiral line. Green: reference trajectories. Blue: real trajectories executed by our system. Red dots: keypoints that POM needs to sequentially go through.}
    \label{trajectory}
    \vspace{-0.3cm}
\end{figure*}

As defined in Sec.~\ref{problem}, we tasked our framework to trace a reference trajectory of POM through in-hand manipulation.
The reference trajectories we used in experiments are shown by the green lines in Fig.~\ref{trajectory}, including a triangle, a square, a $\pi$ letter, and a spiral line.
Each trajectory is represented by a sequence of desired positions of POM (i.e., keypoints, shown by the red dots in Fig.~\ref{trajectory}) in the camera's frame, 
therefore enabling our system to work without the necessity of hand-eye calibration.
To trace a reference trajectory, the POM on the object must reach each keypoint in the correct order, with a tolerance of $\alpha = 1 mm$ in Alg.~\ref{alg:control-system}.
The blue lines in Fig.~\ref{trajectory} showcase POM's actual trajectories while the system executes controls generated by MPC.

Besides the requirement for precise motion control, such an in-hand manipulation task challenges our framework from three other aspects: 
1) With a lack of sensing capability for joint configuration and contact information, our framework needs to be effective in approximating the actual system transitions through self-identification, which is crucial for precise manipulation.
2) The real-world data collection through online manipulation is time-consuming, demanding high data efficiency of the model self-identification. 
3) The contacts between the hand and the object change during manipulation, due to unpredictable sliding or rolling. This requires the adaptability of our framework to such changes.
As will be shown with experiments in Sec.~\ref{ex1}, ~\ref{ex2}, and ~\ref{ex3}, our framework is effective in addressing these challenges and is able to precisely control the object's motion with self-identified models under various real-world settings.

To quantitatively evaluate the performance of our framework on in-hand manipulation, we selected two metrics:
\begin{enumerate}
    \item \emph{Manipulation error}, as defined in Sec.~\ref{resample}, averaged over the entire trajectory. 
    This reflects how far the actual execution deviates from the desired reference trajectory.
    A small manipulation error is a direct indication of accurate motion control of the object, which is highly affected by the quality of the self-identified models and the robustness of our MPC control policy.

    \item \emph{The accumulated number of adapting actions}.
    As described in Sec.~\ref{resample}, whenever the manipulation error exceeds a threshold $\gamma = 2 mm$, our system will perform more exploratory actions to update the self-identified models.
    A small value of this metric means fewer times of model updates, thus reflecting the high data efficiency and good adaptability of our framework.
    
\end{enumerate}
Note that the object was constrained to move in a horizontal plane by our setup. 
The reference trajectories were always given on this plane with the fixed height, and we evaluated the manipulation errors only in this plane as well.


\subsection{Analysis on Initial Exploratory Actions}	
\label{ex1}

In this experiment, we study how many initial exploratory actions are needed for a decent model of self-identification.
With the experiment results, we intend to show that the self-identified models, even learned with only a small amount of training data, can enable precise in-hand manipulation.

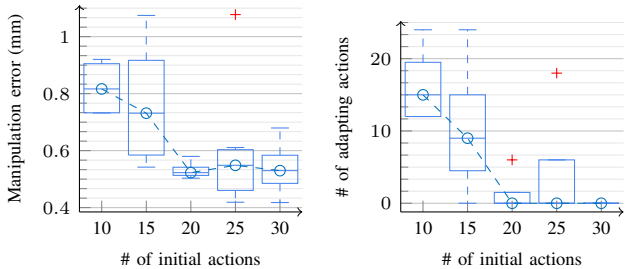
\begin{figure}[!h]
    \vspace{-0.3cm}
    \centering
    \begin{minipage}[b]{0.49\linewidth}
        \definecolor{mycolor1}{rgb}{0.28235,0.52157,0.92941}%
\definecolor{mycolor2}{rgb}{0.00000,0.44700,0.74100}%
\begin{tikzpicture}

\begin{axis}[%
width=0.7\columnwidth,
height=0.65\columnwidth,
scale only axis,
axis x line*=bottom,
axis y line*=left,
xmin=-0.5, 
xmax=4.5,
xtick={0,1,2,3,4},
xticklabels={10,15,20,25,30},
grid=both,
grid style={line width=.1pt, draw=gray!20},
major grid style={line width=.2pt,draw=gray!50},
minor y tick num=5,
xmajorgrids=false,
xminorgrids=false,
ymin=0.384887656125, ymax=1.110473580375,
xlabel={\# of initial actions},
ylabel={Manipulation error (mm)},
]
\addplot [color=mycolor1, dashed, forget plot]
  table[row sep=crcr]{%
0 0.732771594575\\
0 0.7320509861\\
};
\addplot [color=mycolor1, dashed, forget plot]
  table[row sep=crcr]{%
0 0.90512038235\\
0 0.9201957248\\
};
\addplot [color=mycolor1, dashed, forget plot]
  table[row sep=crcr]{%
1 0.58449768165\\
1 0.5420794275\\
};
\addplot [color=mycolor1, dashed, forget plot]
  table[row sep=crcr]{%
1 0.91689757295\\
1 1.074507716\\
};
\addplot [color=mycolor1, dashed, forget plot]
  table[row sep=crcr]{%
2 0.513126178625\\
2 0.503300006\\
};
\addplot [color=mycolor1, dashed, forget plot]
  table[row sep=crcr]{%
2 0.54178497855\\
2 0.5798449098\\
};
\addplot [color=mycolor1, dashed, forget plot]
  table[row sep=crcr]{%
3 0.46047106355\\
3 0.4188732535\\
};
\addplot [color=mycolor1, dashed, forget plot]
  table[row sep=crcr]{%
3 0.603354326575\\
3 0.6109785049\\
};
\addplot [color=mycolor1, dashed, forget plot]
  table[row sep=crcr]{%
4 0.4851248949\\
4 0.4178688345\\
};
\addplot [color=mycolor1, dashed, forget plot]
  table[row sep=crcr]{%
4 0.58392163865\\
4 0.6795251537\\
};

\addplot [color=mycolor1, forget plot]
  table[row sep=crcr]{%
-0.2 0.7320509861\\
0.2 0.7320509861\\
};
\addplot [color=mycolor1, forget plot]
  table[row sep=crcr]{%
-0.2 0.9201957248\\
0.2 0.9201957248\\
};
\addplot [color=mycolor1, forget plot]
  table[row sep=crcr]{%
0.8 0.5420794275\\
1.2 0.5420794275\\
};
\addplot [color=mycolor1, forget plot]
  table[row sep=crcr]{%
0.8 1.074507716\\
1.2 1.074507716\\
};
\addplot [color=mycolor1, forget plot]
  table[row sep=crcr]{%
1.8 0.503300006\\
2.2 0.503300006\\
};
\addplot [color=mycolor1, forget plot]
  table[row sep=crcr]{%
1.8 0.5798449098\\
2.2 0.5798449098\\
};
\addplot [color=mycolor1, forget plot]
  table[row sep=crcr]{%
2.8 0.4188732535\\
3.2 0.4188732535\\
};
\addplot [color=mycolor1, forget plot]
  table[row sep=crcr]{%
2.8 0.6109785049\\
3.2 0.6109785049\\
};
\addplot [color=mycolor1, forget plot]
  table[row sep=crcr]{%
3.8 0.4178688345\\
4.2 0.4178688345\\
};
\addplot [color=mycolor1, forget plot]
  table[row sep=crcr]{%
3.8 0.6795251537\\
4.2 0.6795251537\\
};
\addplot [color=mycolor1, forget plot]
  table[row sep=crcr]{%
-0.4 0.732771594575\\
0.4 0.732771594575\\
0.4 0.90512038235\\
-0.4 0.90512038235\\
-0.4 0.732771594575\\
};
\addplot [color=mycolor1, forget plot]
  table[row sep=crcr]{%
0.6 0.58449768165\\
1.4 0.58449768165\\
1.4 0.91689757295\\
0.6 0.91689757295\\
0.6 0.58449768165\\
};
\addplot [color=mycolor1, forget plot]
  table[row sep=crcr]{%
1.6 0.513126178625\\
2.4 0.513126178625\\
2.4 0.54178497855\\
1.6 0.54178497855\\
1.6 0.513126178625\\
};
\addplot [color=mycolor1, forget plot]
  table[row sep=crcr]{%
2.6 0.46047106355\\
3.4 0.46047106355\\
3.4 0.603354326575\\
2.6 0.603354326575\\
2.6 0.46047106355\\
};

\addplot [color=mycolor1, forget plot]
  table[row sep=crcr]{%
3.6 0.4851248949\\
4.4 0.4851248949\\
4.4 0.58392163865\\
3.6 0.58392163865\\
3.6 0.4851248949\\
};
\addplot [color=mycolor1, forget plot]
  table[row sep=crcr]{%
-0.4 0.8165535328\\
0.4 0.8165535328\\
};
\addplot [color=mycolor1, forget plot]
  table[row sep=crcr]{%
0.6 0.73149897915\\
1.4 0.73149897915\\
};
\addplot [color=mycolor1, forget plot]
  table[row sep=crcr]{%
1.6 0.52274995215\\
2.4 0.52274995215\\
};
\addplot [color=mycolor1, forget plot]
  table[row sep=crcr]{%
2.6 0.548514854\\
3.4 0.548514854\\
};
\addplot [color=mycolor1, forget plot]
  table[row sep=crcr]{%
3.6 0.529798691\\
4.4 0.529798691\\
};

\addplot [color=black, draw=none, mark=+, mark options={solid, red}, forget plot]
  table[row sep=crcr]{%
3 1.077492402\\
};
\addplot [color=mycolor2, dashed, mark=o, mark options={solid, mycolor2}]
  table[row sep=crcr]{%
0	0.8165535328\\
1	0.73149897915\\
2	0.52274995215\\
3	0.548514854\\
4	0.529798691\\
};

\end{axis}
\end{tikzpicture}%
    \end{minipage}
     \begin{minipage}[b]{0.49\linewidth}
        \definecolor{mycolor1}{rgb}{0.28235,0.52157,0.92941}%
\definecolor{mycolor2}{rgb}{0.00000,0.44700,0.74100}%
\begin{tikzpicture}

\begin{axis}[%
width=0.7\columnwidth,
height=0.6\columnwidth,
scale only axis,
axis x line*=bottom,
axis y line*=left,
xmin=-0.5, 
xmax=4.5,
xtick={0,1,2,3,4},
xticklabels={10,15,20,25,30},
grid=both,
grid style={line width=.1pt, draw=gray!20},
major grid style={line width=.2pt,draw=gray!50},
minor y tick num=5,
xmajorgrids=false,
xminorgrids=false,
ymin=-1.2, ymax=25.2,
xlabel={\# of initial actions},
ylabel={\# of adapting actions},
]
\addplot [color=mycolor1, dashed, forget plot]
  table[row sep=crcr]{%
0 12\\
0 12\\
};
\addplot [color=mycolor1, dashed, forget plot]
  table[row sep=crcr]{%
0 19.5\\
0 24\\
};
\addplot [color=mycolor1, dashed, forget plot]
  table[row sep=crcr]{%
1 4.5\\
1 0\\
};
\addplot [color=mycolor1, dashed, forget plot]
  table[row sep=crcr]{%
1 15\\
1 24\\
};
\addplot [color=mycolor1, dashed, forget plot]
  table[row sep=crcr]{%
2 0\\
2 0\\
};
\addplot [color=mycolor1, dashed, forget plot]
  table[row sep=crcr]{%
2 1.5\\
2 1.5\\
};
\addplot [color=mycolor1, dashed, forget plot]
  table[row sep=crcr]{%
3 0\\
3 0\\
};
\addplot [color=mycolor1, dashed, forget plot]
  table[row sep=crcr]{%
3 6\\
3 6\\
};
\addplot [color=mycolor1, dashed, forget plot]
  table[row sep=crcr]{%
4 0\\
4 0\\
};
\addplot [color=mycolor1, dashed, forget plot]
  table[row sep=crcr]{%
4 0\\
4 0\\
};

\addplot [color=mycolor1, forget plot]
  table[row sep=crcr]{%
-0.2 12\\
0.2 12\\
};
\addplot [color=mycolor1, forget plot]
  table[row sep=crcr]{%
-0.2 24\\
0.2 24\\
};
\addplot [color=mycolor1, forget plot]
  table[row sep=crcr]{%
0.8 0\\
1.2 0\\
};
\addplot [color=mycolor1, forget plot]
  table[row sep=crcr]{%
0.8 24\\
1.2 24\\
};
\addplot [color=mycolor1, forget plot]
  table[row sep=crcr]{%
1.8 0\\
2.2 0\\
};
\addplot [color=mycolor1, forget plot]
  table[row sep=crcr]{%
1.8 1.5\\
2.2 1.5\\
};
\addplot [color=mycolor1, forget plot]
  table[row sep=crcr]{%
2.8 0\\
3.2 0\\
};
\addplot [color=mycolor1, forget plot]
  table[row sep=crcr]{%
2.8 6\\
3.2 6\\
};
\addplot [color=mycolor1, forget plot]
  table[row sep=crcr]{%
3.8 0\\
4.2 0\\
};
\addplot [color=mycolor1, forget plot]
  table[row sep=crcr]{%
3.8 0\\
4.2 0\\
};
\addplot [color=mycolor1, forget plot]
  table[row sep=crcr]{%
-0.4 12\\
0.4 12\\
0.4 19.5\\
-0.4 19.5\\
-0.4 12\\
};
\addplot [color=mycolor1, forget plot]
  table[row sep=crcr]{%
0.6 4.5\\
1.4 4.5\\
1.4 15\\
0.6 15\\
0.6 4.5\\
};
\addplot [color=mycolor1, forget plot]
  table[row sep=crcr]{%
1.6 0\\
2.4 0\\
2.4 1.5\\
1.6 1.5\\
1.6 0\\
};
\addplot [color=mycolor1, forget plot]
  table[row sep=crcr]{%
2.6 0\\
3.4 0\\
3.4 6\\
2.6 6\\
2.6 0\\
};

\addplot [color=mycolor1, forget plot]
  table[row sep=crcr]{%
3.6 0\\
4.4 0\\
4.4 0\\
3.6 0\\
3.6 0\\
};
\addplot [color=mycolor1, forget plot]
  table[row sep=crcr]{%
-0.4 15\\
0.4 15\\
};
\addplot [color=mycolor1, forget plot]
  table[row sep=crcr]{%
0.6 9\\
1.4 9\\
};
\addplot [color=mycolor1, forget plot]
  table[row sep=crcr]{%
1.6 0\\
2.4 0\\
};
\addplot [color=mycolor1, forget plot]
  table[row sep=crcr]{%
2.6 0\\
3.4 0\\
};
\addplot [color=mycolor1, forget plot]
  table[row sep=crcr]{%
3.6 0\\
4.4 0\\
};
\addplot [color=black, draw=none, mark=+, mark options={solid, red}, forget plot]
  table[row sep=crcr]{%
2 6\\
};

\addplot [color=black, draw=none, mark=+, mark options={solid, red}, forget plot]
  table[row sep=crcr]{%
3 18\\
};
\addplot [color=mycolor2, dashed, mark=o, mark options={solid, mycolor2}]
  table[row sep=crcr]{%
0	15\\
1	9\\
2	0\\
3	0\\
4	0\\
};

\end{axis}
\end{tikzpicture}%
    \end{minipage}
    \caption{Self-identification performance in terms of the number of initial exploratory actions. For different numbers of initial actions, the result is averaged over all the objects and reference trajectories.}
    \label{fig:training_manipulation}
    \vspace{-0.3cm}
\end{figure}

For this, we varied the number of initial exploratory actions (i.e., $d + a$ in Alg.~\ref{alg:control-system}) to be $10, 15, 20, 25$ and $30$, and tasked the framework to manipulate all the objects in Fig.~\ref{object} and trace all the four trajectories given in Fig.~\ref{trajectory}.
For each setting, we repeated the experiment $5$ times.
To ensure the manipulation performance is not dominated by a bad control policy, we set the MPC optimization scale not too small with $\sigma=0.1$.
The results are summarized in Fig.~\ref{fig:training_manipulation}.
From the results, we find fewer initial exploratory actions cause a worse quality of the self-identified system models reflected by a higher manipulation error and demand for more adapting actions.
This is because the underlying system transitions can be hardly approximated via self-identification, if without sufficient exploration.
However, by slightly increasing the number of initial exploratory actions, higher manipulation precision was achieved with lower manipulation errors; and the non-parametric models were better approximated via self-identification, indicated by a smaller number of adapting actions for the model update. 
After $20$ initial exploratory actions, the manipulation performance of our framework roughly converged, and an average manipulation error of less than $0.8 mm$ could be achieved for all the objects and reference trajectories.
Importantly, this has demonstrated the high data efficiency of our framework, which in general only needs less than $30$ data points to achieve dexterous in-hand manipulation with high precision.

\subsection{MPC Performance Analysis}
\label{ex2}


The self-identified models $\Gamma$ and $\Gamma^{-1}$ by our framework are locally approximated, thereby can never be perfect due to the lack of prior knowledge about the system and the limited sensing capability.
However, for precise manipulation, we require the controls generated by MPC to be robust enough against imperfect model self-identification.
In this experiment, 
we evaluate the robustness of the self-identified models when the control loop is closed by MPC,
and analyze how it is affected by the optimization scale $\sigma$.

Intuitively, a small $\sigma$ will enforce the MPC policy to be more confident about the self-identified models, thus becoming more greedy but potentially not robust to imperfect self-identifications; whereas a large $\sigma$ will increase the search space of MPC for the optimal control.

\begin{figure}[htbp]
    \vspace{-0.3cm}
    \centering
    \begin{minipage}[b]{0.95\linewidth}
\definecolor{mycolor1}{rgb}{0.95686,0.76078,0.05098}%
\begin{tikzpicture}

\begin{axis}[%
width=0.85\columnwidth,
height=0.35\columnwidth,
scale only axis,
axis x line*=bottom,
axis y line*=left,
xmin=0.2,
xmax=6.8,
grid=both,
grid style={line width=.1pt, draw=gray!20},
major grid style={line width=.2pt,draw=gray!50},
minor y tick num=5,
xmajorgrids=false,
xminorgrids=false,
xtick={1,2,3,4,5,6},
xticklabels={{0.02},{0.03},{0.05},{0.1},{0.2},{0.3}},
ymin=0,
ymax=20,
xlabel={MPC optimization scale ($\sigma$)},
ylabel={\# of adapting actions},
]
\addplot[ybar, bar width=0.8, fill=mycolor1, draw=mycolor1, area legend] table[row sep=crcr] {%
1   12.00\\
2    3.00\\
3    1.50\\
4    3.75\\
5    3.00\\
6    0.75\\
};
\addplot[forget plot, color=white!15!black] table[row sep=crcr] {%
-0.2	0\\
7.2	0\\
};

\addplot [color=black, draw=none]
 plot [error bars/.cd, y dir = both, y explicit]
 table[row sep=crcr, y error plus index=2, y error minus index=3]{%
1   12.00  5.43796192 5.43796192\\
2    3.00   2.99999987 2.99999987\\
3    1.50   1.06066017 1.06066017\\
4    3.75   2.24999999 2.24999999\\
5    3.00   2.12132034 2.12132034\\
6    0.75   0.74999988 0.74999988\\
};

\addplot[color=black, dashed, mark=o, mark options={solid, black}] table[row sep=crcr]{%
1   12.00\\
2    3.00\\
3    1.50\\
4    3.75\\
5    3.00\\
6    0.75\\
};

\end{axis}
\end{tikzpicture}%
    \end{minipage}
    \caption{MPC performance evaluation in terms of its optimization scale $\sigma$ and the number of adapting actions.}
    \label{sigma}
    \vspace{-0.3cm}
\end{figure}
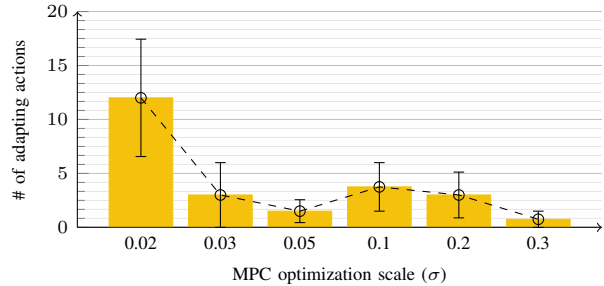

In this specific experiment,
we only used an orange cube (obj \#4) in Fig.~\ref{object} as the object for manipulation.
For MPC, the maximum prediction horizon was set to $K = 5$, and the number of simulated trajectories for optimization was set to $Q = 50$.
For each different $\sigma$, we repeated the manipulation $5$ times for each of the four reference trajectories in Fig.~\ref{trajectory}.
The results are summarized in Fig.~\ref{sigma}.
From the results, we observed a large number of adapting actions with a small $\sigma$ less than $0.02$.
This is expected since MPC is too greedy using the self-identified models with a small $\sigma$, resulting in fewer optimal executions and more instances of model update requests.
By slightly increasing $\sigma$ to introduce more stochasticity into the predictions, 
MPC could search more extensively for finding the optimal control and the average number of adapting actions was immediately reduced to less than $5$, rendering a significant performance improvement.
In general, from the experiments, we found that the self-identified models are sufficiently reliable to make predictions about control with appropriate $\sigma$;
furthermore, our MPC-based control policy was able to achieve precise manipulation by closing the control loop with approximated non-parametric models.

\subsection{Model Generalizability}
\label{ex3}
In real-world applications, the object being manipulated and the grasp configuration are likely to be different every time.
Therefore, good generalizability of the self-identified models to such variations is desirable, as it helps save the time and cost of retraining the models every time.

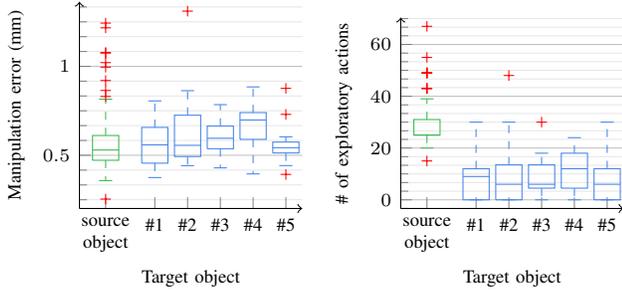
\begin{figure}[!h]
    \vspace{-0.3cm}
    \centering
    \begin{minipage}[b]{0.49\linewidth}
        \definecolor{mycolor1}{rgb}{0.28235,0.52157,0.92941}%
\definecolor{mycolor2}{rgb}{0.23529,0.72941,0.32941}%
\begin{tikzpicture}

\begin{axis}[%
width=0.7\columnwidth,
height=0.65\columnwidth,
scale only axis,
axis x line*=bottom,
axis y line*=left,
xmin=-0.8, 
xmax=6.0,
xtick={0,1.5,2.5,3.5,4.5,5.5},
xticklabels={source\\object, \#1, \#2, \#3,\#4,\#5},
xticklabel style={align=center},
grid=both,
grid style={line width=.1pt, draw=gray!20},
major grid style={line width=.2pt,draw=gray!50},
minor y tick num=5,
xmajorgrids=false,
xminorgrids=false,
ymin=0.20086160013, ymax=1.36299788247,
xlabel={Target object},
ylabel={Manipulation error (mm)},
]
\addplot [color=mycolor2, dashed, forget plot]
  table[row sep=crcr]{%
0 0.471851394175\\
0 0.3573126265\\
};
\addplot [color=mycolor2, dashed, forget plot]
  table[row sep=crcr]{%
0 0.61104167635\\
0 0.8154055751\\
};
\addplot [color=mycolor1, dashed, forget plot]
  table[row sep=crcr]{%
1.5 0.455601154575\\
1.5 0.3747850621\\
};
\addplot [color=mycolor1, dashed, forget plot]
  table[row sep=crcr]{%
1.5 0.6567358998\\
1.5 0.8056234503\\
};
\addplot [color=mycolor1, dashed, forget plot]
  table[row sep=crcr]{%
2.5 0.4928785933\\
2.5 0.4415361021\\
};
\addplot [color=mycolor1, dashed, forget plot]
  table[row sep=crcr]{%
2.5 0.7250830479\\
2.5 0.8630953143\\
};
\addplot [color=mycolor1, dashed, forget plot]
  table[row sep=crcr]{%
3.5 0.5366019438\\
3.5 0.429196349\\
};
\addplot [color=mycolor1, dashed, forget plot]
  table[row sep=crcr]{%
3.5 0.66386562045\\
3.5 0.7845237483\\
};
\addplot [color=mycolor1, dashed, forget plot]
  table[row sep=crcr]{%
4.5 0.58940951465\\
4.5 0.3952234688\\
};
\addplot [color=mycolor1, dashed, forget plot]
  table[row sep=crcr]{%
4.5 0.73992894865\\
4.5 0.884298699\\
};
\addplot [color=mycolor1, dashed, forget plot]
  table[row sep=crcr]{%
5.5 0.512937837675\\
5.5 0.4419433979\\
};
\addplot [color=mycolor1, dashed, forget plot]
  table[row sep=crcr]{%
5.5 0.5742869099\\
5.5 0.6035602478\\
};

\addplot [color=mycolor2, forget plot]
  table[row sep=crcr]{%
-0.2 0.3573126265\\
0.2 0.3573126265\\
};
\addplot [color=mycolor2, forget plot]
  table[row sep=crcr]{%
-0.2 0.8154055751\\
0.2 0.8154055751\\
};

\addplot [color=mycolor1, forget plot]
  table[row sep=crcr]{%
1.3 0.3747850621\\
1.7 0.3747850621\\
};
\addplot [color=mycolor1, forget plot]
  table[row sep=crcr]{%
1.3 0.8056234503\\
1.7 0.8056234503\\
};
\addplot [color=mycolor1, forget plot]
  table[row sep=crcr]{%
2.3 0.4415361021\\
2.7 0.4415361021\\
};
\addplot [color=mycolor1, forget plot]
  table[row sep=crcr]{%
2.3 0.8630953143\\
2.7 0.8630953143\\
};
\addplot [color=mycolor1, forget plot]
  table[row sep=crcr]{%
3.3 0.429196349\\
3.7 0.429196349\\
};
\addplot [color=mycolor1, forget plot]
  table[row sep=crcr]{%
3.3 0.7845237483\\
3.7 0.7845237483\\
};
\addplot [color=mycolor1, forget plot]
  table[row sep=crcr]{%
4.3 0.3952234688\\
4.7 0.3952234688\\
};
\addplot [color=mycolor1, forget plot]
  table[row sep=crcr]{%
4.3 0.884298699\\
4.7 0.884298699\\
};
\addplot [color=mycolor1, forget plot]
  table[row sep=crcr]{%
5.3 0.4419433979\\
5.7 0.4419433979\\
};
\addplot [color=mycolor1, forget plot]
  table[row sep=crcr]{%
5.3 0.6035602478\\
5.7 0.6035602478\\
};

\addplot [color=mycolor2, forget plot]
  table[row sep=crcr]{%
-0.4 0.471851394175\\
0.4 0.471851394175\\
0.4 0.61104167635\\
-0.4 0.61104167635\\
-0.4 0.471851394175\\
};
\addplot [color=mycolor1, forget plot]
  table[row sep=crcr]{%
1.1 0.455601154575\\
1.9 0.455601154575\\
1.9 0.6567358998\\
1.1 0.6567358998\\
1.1 0.455601154575\\
};
\addplot [color=mycolor1, forget plot]
  table[row sep=crcr]{%
2.1 0.4928785933\\
2.9 0.4928785933\\
2.9 0.7250830479\\
2.1 0.7250830479\\
2.1 0.4928785933\\
};
\addplot [color=mycolor1, forget plot]
  table[row sep=crcr]{%
3.1 0.5366019438\\
3.9 0.5366019438\\
3.9 0.66386562045\\
3.1 0.66386562045\\
3.1 0.5366019438\\
};

\addplot [color=mycolor1, forget plot]
  table[row sep=crcr]{%
4.1 0.58940951465\\
4.9 0.58940951465\\
4.9 0.73992894865\\
4.1 0.73992894865\\
4.1 0.58940951465\\
};

\addplot [color=mycolor1, forget plot]
  table[row sep=crcr]{%
5.1 0.512937837675\\
5.9 0.512937837675\\
5.9 0.5742869099\\
5.1 0.5742869099\\
5.1 0.512937837675\\
};

\addplot [color=mycolor2, forget plot]
  table[row sep=crcr]{%
-0.4 0.52962340625\\
0.4 0.52962340625\\
};
\addplot [color=mycolor1, forget plot]
  table[row sep=crcr]{%
1.1 0.558441414\\
1.9 0.558441414\\
};
\addplot [color=mycolor1, forget plot]
  table[row sep=crcr]{%
2.1 0.55588556695\\
2.9 0.55588556695\\
};
\addplot [color=mycolor1, forget plot]
  table[row sep=crcr]{%
3.1 0.5954708252\\
3.9 0.5954708252\\
};
\addplot [color=mycolor1, forget plot]
  table[row sep=crcr]{%
4.1 0.69820231675\\
4.9 0.69820231675\\
};
\addplot [color=mycolor1, forget plot]
  table[row sep=crcr]{%
5.1 0.54232604475\\
5.9 0.54232604475\\
};
\addplot [color=black, draw=none, mark=+, mark options={solid, red}, forget plot]
  table[row sep=crcr]{%
0 0.2536859766\\
0 0.8643608586\\
0 1.074507716\\
0 1.077492402\\
0 0.9201957248\\
0 0.9952682\\
0 1.216297623\\
0 0.8305342243\\
0 1.242825221\\
0 1.02092923\\
};
\addplot [color=black, draw=none, mark=+, mark options={solid, red}, forget plot]
  table[row sep=crcr]{%
2.5 1.310173506\\
};
\addplot [color=black, draw=none, mark=+, mark options={solid, red}, forget plot]
  table[row sep=crcr]{%
5.5 0.3925087213\\
5.5 0.7298981464\\
5.5 0.876250712\\
};

\end{axis}
\end{tikzpicture}%
    \end{minipage}
     \begin{minipage}[b]{0.49\linewidth}
        \definecolor{mycolor1}{rgb}{0.28235,0.52157,0.92941}%
\definecolor{mycolor2}{rgb}{0.23529,0.72941,0.32941}%
\begin{tikzpicture}

\begin{axis}[%
width=0.7\columnwidth,
height=0.6\columnwidth,
scale only axis,
axis x line*=bottom,
axis y line*=left,
xmin=-0.8, 
xmax=6.0,
xtick={0,1.5,2.5,3.5,4.5,5.5},
xticklabels={source\\object, \#1, \#2, \#3,\#4,\#5},
xticklabel style={align=center},
grid=both,
grid style={line width=.1pt, draw=gray!20},
major grid style={line width=.2pt,draw=gray!50},
minor y tick num=5,
xmajorgrids=false,
xminorgrids=false,
ymin=-3.35, ymax=70.35,
xlabel={Target object},
ylabel={\# of exploratory actions},
]
\addplot [color=mycolor2, dashed, forget plot]
  table[row sep=crcr]{%
0 25\\
0 20\\
};
\addplot [color=mycolor2, dashed, forget plot]
  table[row sep=crcr]{%
0 31\\
0 39\\
};
\addplot [color=mycolor1, dashed, forget plot]
  table[row sep=crcr]{%
1.5 0\\
1.5 0\\
};
\addplot [color=mycolor1, dashed, forget plot]
  table[row sep=crcr]{%
1.5 12\\
1.5 30\\
};
\addplot [color=mycolor1, dashed, forget plot]
  table[row sep=crcr]{%
2.5 0\\
2.5 0\\
};
\addplot [color=mycolor1, dashed, forget plot]
  table[row sep=crcr]{%
2.5 13.5\\
2.5 30\\
};
\addplot [color=mycolor1, dashed, forget plot]
  table[row sep=crcr]{%
3.5 4.5\\
3.5 0\\
};
\addplot [color=mycolor1, dashed, forget plot]
  table[row sep=crcr]{%
3.5 13.5\\
3.5 18\\
};
\addplot [color=mycolor1, dashed, forget plot]
  table[row sep=crcr]{%
4.5 4.5\\
4.5 0\\
};
\addplot [color=mycolor1, dashed, forget plot]
  table[row sep=crcr]{%
4.5 18\\
4.5 24\\
};

\addplot [color=mycolor1, dashed, forget plot]
  table[row sep=crcr]{%
5.5 0\\
5.5 0\\
};

\addplot [color=mycolor1, dashed, forget plot]
  table[row sep=crcr]{%
5.5 12\\
5.5 30\\
};

\addplot [color=mycolor2, forget plot]
  table[row sep=crcr]{%
-0.2 20\\
0.2 20\\
};
\addplot [color=mycolor2, forget plot]
  table[row sep=crcr]{%
-0.2 39\\
0.2 39\\
};

\addplot [color=mycolor1, forget plot]
  table[row sep=crcr]{%
1.3 0\\
1.7 0\\
};
\addplot [color=mycolor1, forget plot]
  table[row sep=crcr]{%
1.3 30\\
1.7 30\\
};
\addplot [color=mycolor1, forget plot]
  table[row sep=crcr]{%
2.3 0\\
2.7 0\\
};
\addplot [color=mycolor1, forget plot]
  table[row sep=crcr]{%
2.3 30\\
2.7 30\\
};
\addplot [color=mycolor1, forget plot]
  table[row sep=crcr]{%
3.3 0\\
3.7 0\\
};
\addplot [color=mycolor1, forget plot]
  table[row sep=crcr]{%
3.3 18\\
3.7 18\\
};
\addplot [color=mycolor1, forget plot]
  table[row sep=crcr]{%
4.3 0\\
4.7 0\\
};
\addplot [color=mycolor1, forget plot]
  table[row sep=crcr]{%
4.3 24\\
4.7 24\\
};
\addplot [color=mycolor1, forget plot]
  table[row sep=crcr]{%
5.3 0\\
5.7 0\\
};
\addplot [color=mycolor1, forget plot]
  table[row sep=crcr]{%
5.3 30\\
5.7 30\\
};

\addplot [color=mycolor2, forget plot]
  table[row sep=crcr]{%
-0.4 25\\
0.4 25\\
0.4 31\\
-0.4 31\\
-0.4 25\\
};
\addplot [color=mycolor1, forget plot]
  table[row sep=crcr]{%
1.1 0\\
1.9 0\\
1.9 12\\
1.1 12\\
1.1 0\\
};
\addplot [color=mycolor1, forget plot]
  table[row sep=crcr]{%
2.1 0\\
2.9 0\\
2.9 13.5\\
2.1 13.5\\
2.1 0\\
};
\addplot [color=mycolor1, forget plot]
  table[row sep=crcr]{%
3.1 4.5\\
3.9 4.5\\
3.9 13.5\\
3.1 13.5\\
3.1 4.5\\
};

\addplot [color=mycolor1, forget plot]
  table[row sep=crcr]{%
4.1 4.5\\
4.9 4.5\\
4.9 18\\
4.1 18\\
4.1 4.5\\
};

\addplot [color=mycolor1, forget plot]
  table[row sep=crcr]{%
5.1 0\\
5.9 0\\
5.9 12\\
5.1 12\\
5.1 0\\
};

\addplot [color=mycolor2, forget plot]
  table[row sep=crcr]{%
-0.4 25\\
0.4 25\\
};
\addplot [color=mycolor1, forget plot]
  table[row sep=crcr]{%
1.1 9\\
1.9 9\\
};
\addplot [color=mycolor1, forget plot]
  table[row sep=crcr]{%
2.1 6\\
2.9 6\\
};
\addplot [color=mycolor1, forget plot]
  table[row sep=crcr]{%
3.1 6\\
3.9 6\\
};
\addplot [color=mycolor1, forget plot]
  table[row sep=crcr]{%
4.1 12\\
4.9 12\\
};
\addplot [color=mycolor1, forget plot]
  table[row sep=crcr]{%
5.1 6\\
5.9 6\\
};
\addplot [color=black, draw=none, mark=+, mark options={solid, red}, forget plot]
  table[row sep=crcr]{%
0 15\\
0 43\\
0 43\\
0 43\\
0 67\\
0 49\\
0 49\\
0 55\\
0 49\\
0 49\\
};
\addplot [color=black, draw=none, mark=+, mark options={solid, red}, forget plot]
  table[row sep=crcr]{%
2.5 48\\
};
\addplot [color=black, draw=none, mark=+, mark options={solid, red}, forget plot]
  table[row sep=crcr]{%
3.5 30\\
};

\end{axis}
\end{tikzpicture}%
    \end{minipage}
    \caption{Results of model transfer experiments. The model was self-identified on a source object (green) and then used for manipulating different other target objects (blue).}
    \label{fig:transfer}
    \vspace{-0.3cm}
\end{figure}

In this experiment, we challenged the generalizability of the self-identified models in our framework, by manipulating a new object (\emph{target object}) with models learned through manipulating a different object (\emph{source object}). 
Specifically, we saved the models $\Gamma$ and $\Gamma^{-1}$ learned from manipulating an orange cube (obj \#4) with $25$ initial exploratory actions.
Similar to Sec.~\ref{sec:transfer}, we directly used this saved model to initialize the manipulation of a new object, without any initial exploratory actions on the new object.
For each object, we ran $4$ trials for each reference trajectory and averaged the results in Fig.~\ref{fig:transfer}.
As can be seen from the results, regardless of the specific object, there was no significant increase in manipulation errors even though the manipulation models were learned using a different object.
When manipulating a new object, although we observed that more adapting actions were performed for the model update, the total number of exploratory actions was reduced from $25$ to $12$ since no exploratory actions were needed initially.
In summary, the experiment has shown good generalizability of the self-identified models, and efficient utilization of model transfer.


\section{Conclusion}
\label{conclusion}
In this paper, we approached the in-hand dexterous manipulation problem with non-parametric self-identification and Model Predictive Control.
With a small number of exploratory actions, our proposed framework efficiently self-identifies the underlying manipulation models through Gaussian Process Regression. 
By integrating the self-identified manipulation models into an MPC-based framework, a robust control method can be developed for precise manipulation.
Furthermore, when the self-identified local models become unreliable for generating effective controls, our framework can adaptively update the models by performing more exploratory actions.

With extensive real-world experiments on an underactuated Yale Model O hand, we show that 1) our proposed framework can achieve millimeter-level in-hand manipulation without requiring a large amount of data nor sophisticated sensing systems; 2) the MPC-based control is robust when integrated with imperfect manipulation models, which are locally approximated; 3) the self-identified manipulation models can well generalize on similar setup variations.


In future work, we aim to implement the proposed framework on a more complex or higher-dimensional system, for example, an underactuated hand with more motors or a compliant robot arm with higher degrees of freedom.
Furthermore, we plan to improve the current framework, to achieve higher precision and dexterity of in-hand manipulation by a more capable control policy; or to enhance the adaptability of self-identification to different types of motion constraints.


\bibliographystyle{IEEEtran}
\bibliography{reference}

\end{document}